\documentclass[11pt]{article}

\usepackage{jmlr2e}

\usepackage[dvipsnames]{xcolor}
\usepackage{amsfonts} 
\usepackage{graphicx}
\usepackage{amsfonts}
\usepackage{algorithm}
\usepackage{algorithmicx}
\usepackage{algpseudocode}
\usepackage{array}
\usepackage{amsmath}
\usepackage{booktabs}
\usepackage{siunitx}
\usepackage{tikz}

\newcommand{\balpha}{\boldsymbol\alpha}

\newcommand{\btheta}{\boldsymbol\theta}
\newcommand{\bTheta}{\boldsymbol\Theta}

\newcommand{\bmu}{\boldsymbol\mu}
\newcommand{\bgamma}{\boldsymbol\gamma}
\newcommand{\bGamma}{\boldsymbol\Gamma}

\newcommand{\bX}{\boldsymbol X}

\newcommand{\bx}{\boldsymbol x}
\newcommand{\by}{\boldsymbol y}
\usepackage{url}

\usepackage[dvipsnames]{xcolor}

\long\def\comment#1{}

\jmlrheading{1}{2023}{1-xx}{5/23}{xx/23}{Hlav\'a\v ckov\'a-Schindler, Melnykova and Tubikanec}

\ShortHeadings{}{}
\firstpageno{1}

\begin{document}

\title{Granger Causal Inference in Multivariate Hawkes Processes 
by Minimum Message Length}

\author{\name Kate\v{r}ina Hlav\'a\v{c}kov\'{a}-Schindler \email katerina.schindlerova@univie.ac.at \\
       \addr
       Faculty of Computer Science, University of Vienna \\
       Vienna, Austria, and Institute of Computer Science, \\Czech Academy of Sciences, Prague, Czechia
       \AND
       \name Anna Melnykova \email anna.melnykova@univ-avignon.fr \\
       \addr  Laboratory of Mathematics\\
       University of Avignon, Avignon, France
       \AND
       \name Irene Tubikanec \email irene.tubikanec@aau.at \\
       \addr Department of Statistics\\
       University of Klagenfurt, Klagenfurt, Austria}

\editor{}

\maketitle

\vspace{0.5cm}

\begin{abstract}%

Multivariate Hawkes processes (MHPs) are versatile probabilistic tools used to model various real-life phenomena: earthquakes, operations on stock markets, neuronal activity, virus propagation and many others. In this paper, we focus on MHPs with exponential decay kernels and estimate connectivity graphs, which represent the Granger causal relations between their components. We approach this inference problem by proposing an optimization criterion and model selection algorithm based on the minimum message length (MML) principle. MML compares Granger causal models using the Occam's razor principle in the following way: even when  models  have a comparable goodness-of-fit to the observed data, the one generating the most concise explanation of the data is preferred. While most of the state-of-art methods using lasso-type penalization tend to overfitting in scenarios with short time horizons, the proposed MML-based method achieves high F1 scores in these settings. We conduct a numerical study comparing the proposed algorithm to other related classical and state-of-art methods, where we achieve the highest F1 scores in specific sparse graph settings. We illustrate the proposed method also on G7 sovereign bond data and obtain causal connections, which are in agreement with the expert knowledge available in the literature. 
\end{abstract}


\section{Introduction}\label{section:intro}

Many practical applications  deal with a large
amount of irregular and asynchronous sequential data observed within a fixed time  horizon. 
One can interpret such data as event sequences containing stereotypic events,  
which can be modeled via multidimensional point processes. 
These events can be, e.g. user viewing records, patient records in hospitals (in which times,  diagnoses or treatments are provided), various levels of earthquakes, high-frequency financial transactions or neuronal activity.  

In this paper, we focus on a special type of point processes, known as Hawkes processes \citep{hawkes1971spectra}.
Their main advantage over other point processes (such as the classical Poisson processes) is that they permit to model the influence of past events, thanks to their ``memory'' property, as well as possible interactions between different components of the process. 
In particular, a multivariate Hawkes process (MHP) $\bX=(\bX_t)_{t\in[0,T]}=(X_t^1,\ldots,X_t^p)_{t \in [0,T]}$ is a $p$-dimensional temporal point process representing a system of $p\geq 2$ interacting units. One can interpret each component of a MHP as a ``particle'' or ``node'' in some given system, e.g. a neuron in a brain, an account in a social network, or a certain type of financial transaction. Here, we consider MHPs with conditional intensity function, at each dimension $i\in \{ 1,\ldots,p \}$ following
\begin{equation}\label{eq:intensity_expMHP}
    \lambda_i(t) = \mu_i + \sum_{j=1}^p 
    \int_0^{t} \alpha_{ij}\exp( -\beta_{ij}(t-\tau))d
    X^j_{\tau},
\end{equation}
where the $\mu_i>0$ are positive  parameters also known as background intensities, $\beta_{ij}>0$ are positive 
decay constants, and $\alpha_{ij}\geq 0$ are non-negative influence parameters, which model the interaction between different components of the process $\bX$. The conditional intensity function gives an expected number of events on each infinitely small interval of time, i.e.
\begin{equation}
    \lambda_i(t) = \mathbb{E}[d{X}^i_t|{\cal F}_t] =  \lim_{\Delta t \to 0} \mathbb{E}[{X}^i_{t + \Delta t} - {X}^i_{t}| {\cal F}_t],
\end{equation}
where ${\cal F}_t$ is a filtration which contains all the information of the process prior to time~$t$. A realization of the process corresponds to a list of event occurrence times within the time interval $[0,T]$ at which the counts are carried out. In particular, for each $i\in \{ 1,\ldots,p \}$, the vector of observed event times of the $i$-th particle $(X^i_t)_{t\in [0,T]}$ is given by $\bx_i:=(t^i_1, \dots, t^i_{n_i})^\top$, where $n_i \in \mathbb{N}$ and $0 < t^i_1 < \dots < t^i_{n_i}\leq T$, and a realization of the entire process $\bX$ is given by $\bx=\{ \bx_i \}_{i=1}^{p}$. The value $T$ is called the time horizon of a MHP. The case when $T$ is of order at most a hundred times the dimension $p$ is referred to as a ``short'' time horizon. If $T$ is of order at least a thousands times the dimension $p$, we talk about a ``long'' time horizon.

Since we focus on interaction functions given by an exponential kernel, in the following we will refer to $\bX$ as exp-MHP. The main objective of this paper is to infer the connectivity graph, which describes the Granger-causal relationships between the components of exp-MHPs.
We use the notion of Granger causality among Hawkes processes based on the definition from    \cite{Eichler2015} and say that the component $X^j$ does not \textit{Granger-cause} $X^i$ if and only if the corresponding interaction function 
is equal to $0$ for all $t \in [0,T]$. 
Since our interaction function is given by
    $\alpha_{ij}\exp( -\beta_{ij}(t-\tau))$
this holds if and only if the influence parameter $\alpha_{ij}=0$. In this case, there is no  edge  leading from $j$ to $i$ in the corresponding  graph, otherwise an edge $j\to i$ is present in the graph. In other words, we study the  problem of estimating the connections in a directed graph when the underlying model is an exp-MHP. 

We approach this inference problem by proposing a model selection algorithm called MMLH for exp-MHPs based on the so called minimum message length (MML) principle.
Methods using MML learn through a data compression perspective and are sometimes described as mathematical applications of Occam's razor, see e.g.  \cite{grunwald2019minimum}.
The minimum message length  principle for statistical and inductive inference as well as machine learning was originally introduced 
 by  \cite{wallace3}. It is a formal information-theoretic restatement of Occam's razor: 
This means that even when models have a comparable goodness-of-fit accuracy to the observed data, the one generating the shortest overall message is more likely to be correct. In this context, a message consists of a statement of the model, followed by a statement of the data encoded concisely using that model. The MML method considers the model which compresses the data most (i.e., the one with the ``shortest message length'') as most descriptive for the data.

As the proposed MMLH algorithm to recover causal connections in exp-MHPs is a model selection method, it allows to incorporate possible expert knowledge about the underlying structure.
For example, this may be knowledge about the maximum number $m$ of possible causal connections to each node, such as the maximum number of debtors for every trustee in a financial connectivity graph. If such knowledge is available, the algorithm searches over a reduced set of possible structures (those indicating at most $m$ connections), decreasing the number of parameters which have to be simultaneously estimated under a given structure.
Parametric inference methods on the  contrary, e.g. maximum-likelihood estimation (MLE), require all parameters to be estimated simultaneously, which can result in a poor performance. In contrast to other model selection methods, such as the classical one obtained via the Bayesian information criterion (BIC) or the recent method proposed by \cite{jalaldoust} based on the data compression technique ``minimum description length'' (MDL), MMLH incorporates prior distributions of relevant model parameters, making the method more flexible in terms of structure-related penalty.

We compare the proposed MMLH algorithm to two state-of-the-art methods (the related MDL-based method from \cite{jalaldoust} and the method ADM4 from \cite{zhou2013learningb}), as well as to three standard reference methods, namely BIC, AIC (Akaike information criterion), and MLE. We focus on data with short time horizons and consider graphs of dimension seven, ten, and twenty, respectively. MMLH shows the highest F1 accuracy with respect to all considered methods for specific sparse graph settings. We complete the numerical study by applying our approach on a real-world data base, which describes the return volatility of sovereign bonds of seven large economies (see e.g. \cite{demirer2018estimating, jalaldoust}). Most discovered causal connections are in accordance with the expert knowledge from the literature.

\paragraph{Notations.} Regarding terminology, 
we  use both terms ``causal structure'' and ``connectivity graph'', depending on which is more appropriate in the respective context.
Regarding notation, scalar variables are denoted by regular letters and vectors and matrices by bold letters. Stochastic processes 
are denoted by capital letters
(e.g.  $\bX$) and any realization or point by a lower-case letter (e.g. ${\bx}$). Matrices are denoted by Greek or capital regular letters. Let $\balpha$ be a generic matrix. Then $\balpha_{i}$ denotes the $i$-th row of the matrix $\balpha$ and $\alpha_{ij}$ the $j$-th entry in the $i$-th row. Moreover, $\balpha^{\top}$ and $|\balpha|$ denote the transpose and determinant of the matrix $\balpha$, respectively.

This paper is organized as follows. 
Section~\ref{related work} discusses related work. Section~\ref{section:preliminaries} recalls the general idea of MML as a criterion for model selection and parameter estimation.
In Section~\ref{sec:causality_MML}, we apply the MML approach to exp-MHPs. The proposed algorithm  MMLH is described in Section~\ref{sec:algorithm}. In Section ~\ref{section:numerical_experiments}, we illustrate the performance of the proposed algorithm in comparison to benchmark methods on both synthetic data  as well as real-world data. Section~\ref{section:conclusion} concludes the study and outlines perspectives for possible  
future work. MMLH is coded in \texttt{R}, using the package \texttt{Rcpp} (\cite{Rcpp}). A sample code is available at: \url{https://github.com/IreneTubikanec/MMLH}

\vspace{-0.3cm}
\section{Related Work}\label{related work}

Related work can be categorized into the work on discovery of Granger causal networks in MHPs and on applying compression based methods (such as MML and MDL) to Granger causal inference.

The problem of inferring the Granger causal structure is relatively new in the context of MHPs, however, it has attracted a lot of attention in recent years, see  e.g.   \cite{hansen15}, \cite{xu} and \cite{sulem}.  \cite{didelez} studied causal connectivity graphs for discrete  time events and extended them  to marked point processes. 
 Most of the related work deals with variable selection in sparse causal graphs.
The recovery of the Granger causal structure is directly linked to the problem of  (parametric or nonparametric) estimation of the interaction function, which is studied in the literature, e.g. by \cite{Eichler2015}.
The most common approach to reconstruct the network is to apply maximum likelihood estimation, see e.g. \cite{ogata1988},  \cite{veen}, \cite{juditsky}.
Maximum likelihood estimation reveals favorable theoretical properties and is not computationally expensive. However, it does not lead to good scores in practice, especially on small datasets. Some improvement can be achieved when the estimation is done via confidence intervals (as in  \cite{wang}), however, they are difficult to compute for a general class of models. 
\cite{xu} applied an expectation maximization (EM)  algorithm based on a penalized likelihood objective leading to temporal and group sparsity to infer a Granger graph in MHPs. 

The method ADM4 in \cite{zhou2013learningb} performs variable selection by using lasso and nuclear norm regularization simultaneously
on the parameters to cluster variables as well as to obtain a
sparse connectivity graph. The method NPHC (Achab
et al. 2017) takes a non-parametric approach in learning the
norm of the kernel functions to find the causal connectivity graph. The method uses a moment-matching approach to fit
the second-order and third-order integrated cumulants of the
process.

To infer a causal connectivity graph,
\cite{bacry2015-2}
optimize a least-square based objective function  with lasso and trace
norm of the interaction tensor for the intensity process. 
\cite{trouleau2021}
investigated stability of cumulant-based estimators
for causal inference in MHPs with respect to noise.
\cite{wei2023granger}  recover  a Granger causal graph for Hawkes processes coupled with the so-called ReLU link function; It was tested on long time horizons $T$ and, in comparison to other mentioned methods, 
it considers  both exciting and inhibiting effects.
\cite{ide2021} introduced a  causal learning framework based
on a cardinality-regularized Hawkes process. 
\cite{hansen15}  
use lasso penalization to infer sparse connectivity graphs in MHPs.
Most of the above mentioned  methods using lasso-type penalization demonstrated good performance in scenarios with long time horizons $T$. 
It  is   however known that lasso-type penalization methods often suffer from overfitting in the opposite case of
short time horizons, see e.g. \cite{reid2016}. 
To overcome the drawbacks of these methods,  we approach penalization  based on the MML principle.

MML-based  model selection as an inductive inference method based on  data compression  was first introduced in \cite{wallace3}. Intuitively, the recovery of a connectivity graph  using the MML principle is equivalent to selecting an optimal model for the observed data, where ``optimal" means ``the one which permits to encode the data in a binary string of the shortest length" in terms of coding theory. 
There exist papers on the recovery of Granger connectivity graphs by MML  for processes having distributions from exponential families, see \cite{hlavackova2020, hlavackova2020b}, but to the best of our knowledge, not for MHPs. 
Another compression scheme using Occam's razor in terms of coding representations is the minimum description length (\cite{rissanen1998stochastic}), which was more recently developed in \cite{grunwald2007minimum} and \cite{grunwald2019minimum}. In comparison to MML, the MDL principle does not use any knowledge of priors. 
MDL-based Granger-causal inference has been recently applied to exp-MHPs in \cite{jalaldoust}  and to Gaussian processes in  \cite{hlavackova2020b}.

\section{Minimum Message Length Criterion and Its Approximation}\label{section:preliminaries}

Methods based on the MML principle consider the model which compresses the data the most (i.e., the one with the ``shortest message length"). To be able to decompress this representation of the data, the details of the statistical model used to encode the data must also be a part of the compressed data string. The calculation of the exact message is an NP-hard problem, since it corresponds to the Kolmogorov complexity (see \cite{wallace1999}), which is in general not computable due to the halting problem (see  \cite{ming2008}). However, there exist computable approximations of MML, the most used one is the Wallace–Freeman approximation \citep{wf1987}, which we will use in this paper (see Section \ref{subsec:mml_criteria}). 

Before we recall the general idea  behind MML and outline the aforementioned approximation approach, we define statistical models.
Statistical models are families of probability distributions of the form
\begin{equation}\label{modelM}
M = \{p(\cdot |\btheta) : \btheta \in  \bTheta \},
\end{equation} 
parametrized by
a  set $\bTheta$   (usually a subset of a Euclidean space).
They are represented by families
of probability distributions 
\begin{equation} 
\{M_{\bgamma}: \bgamma \in \bGamma\},
\end{equation}
where $\bGamma$ is a countable set of so-called ``structures'' and, for each structure $\bgamma \in \bGamma$,
\begin{equation}
    M_{\bgamma} = \{p_{\bgamma}(\cdot |\btheta) : \btheta \in \bTheta_{\bgamma} \}
\end{equation}
is a statistical model, parameterized by the space $\bTheta_{\bgamma}$. 

In our setting, the set of structures $\bGamma$ can be interpreted as a countable set of binary vectors, i.e. $\bGamma=\{ 0,1 \}^q$ with $q>0$. Each element $\bgamma \in \bGamma$ is then a $q$-dimensional vector of zeros and ones, where the number of ones is given by $k$, and where $\gamma_j=1$ denotes the presence of the $j$-th variable  in the  subset of  $k$ variables, and $\gamma_j=0$ means that the $j$-th variable is not present.  
The parameters $\btheta \in \bTheta_{\bgamma} \subset \mathbb{R}^k$ then define the "weights" (which can be interpreted as importance measures) with respect to the variables in $\bgamma$. 
In the graph context, a structure set $\bGamma=\{ 0,1 \}^q$ corresponds to a given node, which can have at most $q$ causes.
An element $\gamma_j$, $j \in \{1,\ldots,q\}$, of a structure $\bgamma \in \bGamma$ indicates the presence ($\gamma_j=1$) or absence $(\gamma_j=0)$ of an incoming edge from node $j$ into the given node. For example, for a graph with $q=3$ nodes, the corresponding structure set for each node is given by $\bGamma=\{ 0,1 \}^3=\{ (0,0,0),(1,0,0),(0,1,0),(0,0,1),(1,1,0),(1,0,1),(0,1,1),(1,1,1) \}$. If $\bGamma$ corresponds to node $1$, then the element $\bgamma=(0,1,0) \in \bGamma$ indicates that node $1$ is not self-excitatory (since $\gamma_1=0$), has an incoming edge from node $2$ (since $\gamma_2=1$), and has no incoming edge from node $3$ (since $\gamma_3=0$).



\subsection{Idea Behind the Minimum Message Length Method}  

The MML principle is a formal information theory restatement of Occam’s razor: even when models have a comparable goodness-of-fit to the observed data, the one generating the shortest overall message is more likely to be correct (where the message consists of a statement of the model, followed by a statement of data encoded concisely using that model). Let us describe the idea of the MML method more formally. 

Consider some data $\by =(y_1, \dots, y_n)^{\top}
\in \mathbb{R}^n$ that we would like to send to a  receiver by encoding
it into a message (e.g. a binary string). The key idea in MML inference is to interpret this message as consisting of the following parts: an encoding (called \textit{assertion}) of the model structure $\bgamma \in \bGamma$ and associated parameters $\btheta \in \bTheta_{\bgamma}$, a description (called \textit{detail}) of the data $\by$ using the model $p_{\bgamma}({\by}|\btheta)$  specified in the assertion, and a preamble code describing which structure is used. The total message length of the data $\by$, model structure $\bgamma \in \bGamma$, and parameterization $\btheta \in \bTheta_{\bgamma}$ is then given by
\begin{equation}\label{eq:fullcode2}
    I({\by} ; \btheta ; \bgamma) = I(\btheta; \bgamma) + I({\by}|\btheta; \bgamma) + I(\bgamma),
\end{equation} 
where $I(\btheta; \bgamma)$, $I({\by}|\btheta; \bgamma)$, and $I(\bgamma)$ denote the length of the assertion, detail, and structure preamble code, respectively. 
Equation (\ref{eq:fullcode2}) is also called  refined total message length in the literature.
 
The length of the assertion $I(\btheta;\bgamma)$ is a measure of the model complexity, while the length of the detail $I({\by}|\btheta;\bgamma)$ is a measure of the goodness-of-fit of the model to the data (model capability). Moreover, for $\gamma \in \bGamma=\{0,1\}^q$, the set of all possible structures,  we set
    \begin{equation}\label{eq:codelength}
        I(\bgamma) = \log \dbinom{q}{k} + \log(q +1),
    \end{equation}
as recommended by \cite{roos2009}. MML seeks the model structure and corresponding parameters that minimize this trade off between model complexity and  model capability,~i.e.
\begin{equation}\label{eq:hat_gamma_theta2}
\{\hat{\bgamma}, \hat{\btheta}\} = 
\textrm{argmin}_{\bgamma \in \bGamma, \btheta \in \bTheta_{\bgamma}} \ I({\by};\btheta; \bgamma).
\end{equation}

\subsection{Wallace-Freeman Approximation}\label{subsec:mml_criteria}

In the following, we recall a well-known approximation of the total message length introduced in  \eqref{eq:fullcode2}, originally proposed by \cite{wf1987}. A detailed presentation of this approximation can be also found in Chapter 5 of \cite{wallace}, including a list of required assumptions given in Section 5.1.1. Similar to Bayesian selection methods, this procedure utilizes prior probability distributions for parameters.

According to the Wallace-Freeman approximation, the codelength of data $\by$ for a given model parametrization $\btheta \in \bTheta_{\bgamma}$ under a fixed structure $\gamma \in \bGamma=\{ 0,1\}^q$ (i.e., the detail w.r.t. $\gamma$) is given by
\begin{align}\label{eq:I87}
I({\by} | \btheta ; \bgamma) \approx 
- \log p_{\bgamma}({\by}|\btheta)  + \frac{k}{2}.
\end{align}

Moreover, the codelength of the assertion w.r.t. $\gamma$ is given by
\begin{align}\label{eq:I87_2}
I(\btheta;\gamma) \approx 
 - \log \pi_{\bgamma}(\btheta) + \frac{1}{2} \log|J_{\bgamma}(\btheta)| 
+ \frac{k}{2}\log \kappa_k,
\end{align}
where $\pi_{\bgamma}(\btheta)$ is a prior probability distribution over $\bTheta_\gamma$, 
$J_\gamma(\btheta)$ is the expected Fisher information matrix and $\kappa_k$ is a quantizing lattice constant, which depends on the number of parameters $k$ that is determined by the structure $\gamma$.  

\begin{figure}
    \centering
    \includegraphics[width = 0.6\textwidth]{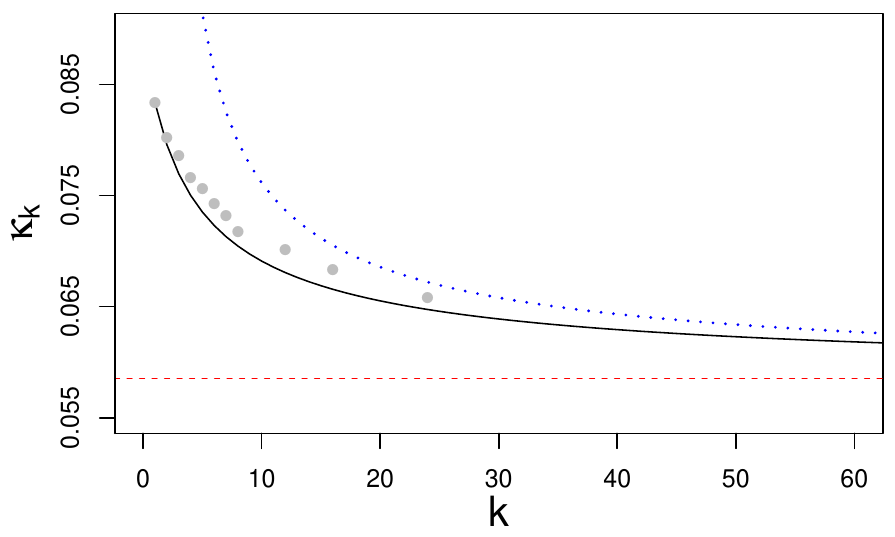}
    \caption{Blue dotted line: Upper bound for $\kappa_k$. Black solid line: Lower bound for $\kappa_k$. Red~dashed line: Limit of the bounds for $k \to \infty$. Grey dots: Known values~of~$\kappa_k$.} 
    \label{fig:bounds}
\end{figure}

While an optimal value for $\kappa_k$ is not available in general, in \cite{wf1987} the following upper and lower bounds were proposed for $k>1$:
\begin{equation}\label{eq:bounds_kappa_k}
    \frac{\Gamma(k/2+1)^{2/k}}{\pi (k+2)} < \kappa_k < \frac{\Gamma(k/2+1)^{2/k} \Gamma(2/k + 1)}{\pi k},
\end{equation}
where in this case  $\Gamma$ denotes the gamma function.
These bounds are reported as function of $k$ (black solid and blue dotted lines) in Figure \ref{fig:bounds}. They both converge to $1/(2\pi e)$ (red dashed line) for $k \to \infty$. Some values of $\kappa_k$ (for small $k$) are known explicitly, see e.g. \cite{conway}, \cite{schmidt2021}. Those reported in Table 1 of \cite{conway} are added as gray dots to Figure \ref{fig:bounds}. For $k=1$, it is known that $\kappa_k=1/12$, and thus the lower bound in \eqref{eq:bounds_kappa_k} is achieved, see~\cite{schmidt2021}. The choice of approximation for $\kappa_k$ influences the penalty term with respect to the number of parameters $k$ determined by the structure $\gamma$. 
Following again \cite{wf1987} and \cite{wallace}, pp. 257--258, we focus on the approximation 
\begin{equation}\label{eq:approx}
\frac{k}{2}( \log \kappa_k + 1) \approx 
-\frac{k}{2} \log(2\pi) + \frac{1}{2} \log(k\pi) + \psi(1),
\end{equation} 
where $\psi$ denotes the digamma function and $\psi(1) \approx  -0.5772$. 

Using \eqref{eq:codelength} and the approximations \eqref{eq:I87}, \eqref{eq:I87_2} and \eqref{eq:approx},  the total message length  
\eqref{eq:fullcode2} 
can be approximated as follows:
\begin{equation}\label{eq:MML87_full}
\begin{aligned}
I(\by;\btheta;\bgamma) 
\approx & - \log p_{\bgamma}({\by}|\btheta)  - \log \pi_{\bgamma}(\btheta) + \frac{1}{2} \log|J_{\bgamma}(\btheta)| \\
&  -\frac{k}{2} \log(2\pi) + \frac{1}{2} \log(k\pi) +  \psi(1) \\
& + \log \dbinom{q}{k} + \log(q +1).
\end{aligned}
\end{equation}

\section{Granger Causal Structure Recovery in Multivariate Hawkes Processes by Minimum Message Length}\label{sec:causality_MML}

In this section, we present the proposed  
MML-based procedure  
for causal inference in exp-MHPs defined via intensity \eqref{eq:intensity_expMHP}. First, we define the corresponding parameter space $\bTheta$. 
Second, we introduce the components required to define the total message length \eqref{eq:MML87_full} for exp-MHPs over the parameter space $\bTheta$. In particular, we report an explicit expression of the  
the log-likelihood, 
derive an approximation for the Fisher information matrix, 
and choose appropriate prior distributions. Finally, we introduce suitable structures, include them in the aforementioned expressions, and propose a criterion for the total message length in exp-MHPs.

\subsection{Parameter Space for Exp-MHPs}\label{section:model}

Consider an exp-MHP $\bX$, i.e. a MHP defined via intensity  (\ref{eq:intensity_expMHP}).
Throughout, we assume for all $i,j \in \{1,\ldots,p\}$ that the decay constants $\beta_{ij}$ are known. This is a common practice in the literature (see, e.g. \cite{juditsky}, \cite{wang}, \cite{jalaldoust}), since these constants are considered to be part of the model itself (such as the memory kernel which is here exponential). Moreover, we assume that the background intensities $\mu_i$  of the $i$-th particle $X^i$ and the influence vector $\mathbf{\balpha}_{i}= (\alpha_{i1}, \dots, \alpha_{ip})^{\top}$ on  $X^i$ are not known. 
Considering the entire process $\bX$, we also introduce the unknown \textit{baseline vector} $\bmu=(\mu_1,\ldots,\mu_p)^{\top}$ and \textit{influence matrix} $\balpha$, whose $i$-th row corresponds to the influence vector $\balpha_i$. Recall that by the definition of Granger causality an entry $\alpha_{ij}$ is  non-zero if and only if there is an incoming edge from node $j$ to node $i$. 
The parameter vector of $\bX$ is then defined   
as
\begin{equation}\label{vectortheta2}
    \btheta = [\btheta_1^{\top}, \btheta_2^{\top}, \dots, \btheta_p^{\top}]^{\top} \in \bTheta = (\mathbb{R}^+_0)^{p+p^2},
\end{equation}
where 
\begin{equation}\label{vectortheta}
    \btheta_i = (\mu_i, \balpha_i^{\top})^{\top} \in \bTheta_i = (\mathbb{R}^+_0)^{p+1}
\end{equation}
is the parameter vector of the $i$-th component $X^i$, for $i \in \{1,\ldots,p\}$.

\subsection{Log-Likelihood for Exp-MHPs}\label{Sec:loglik}

In the following, we recall the log-likelihood of an exp-MHP, see, e.g. \cite{ozaki79} (univariate case) and \cite{Shlomovich2022} (multivariate case).
Consider an observation $\bx$ of an exp-MHP $\bX$ and the parameter vector $\btheta \in \bTheta$ \eqref{vectortheta2}.
Then, the log-likelihood can be decomposed as
\begin{equation}\label{eq:negative_loglik}
\log p (\bx|\btheta) 
= - \sum_{i=1}^p \left( \int_{0}^T\lambda_i(s)ds 
 -  \sum_{j=0}^{n_i} \log \lambda_i(t^i_j) \right).
\end{equation}
Since each summand of this function depends only on the $i$-th dimension $\btheta_i$ \eqref{vectortheta} of the parameter vector $\btheta$ \eqref{vectortheta2}, the negative log-likelihood function can be written as the sum:
\begin{equation}\label{eq:sum:likelihood}
    - \log p(\bx|\btheta) = -\sum_{i=1}^p  \log p_i(\bx|\btheta_i),
\end{equation}
where each summand represents the marginal negative log-likelihood of the corresponding node.
To ease the notations, define $l(\bx|\btheta):=-\log p(\bx|\btheta)$ and $l_i(\bx|\btheta_i):=-\log p_i(\bx|\btheta_i)$.
The explicit expression for each $l_i(\bx|\btheta_i)$ can be derived using \eqref{eq:intensity_expMHP} and is given by
\begin{multline}\label{eq:log_lik}
  l_i(\bx|\btheta_i) = \mu_i 
 t^{\max}
   + \sum_{j=1}^p \frac{\alpha_{ij}}{\beta_{ij}} \sum_{k=1}^{n_j} \left[1-\exp(-\beta_{ij}(
  t^{\max}
  -t^j_k))\right] \\
   - \sum_{l=1}^{n_i} \log\left[ \mu_i + \sum_{j=1}^p \alpha_{ij} \sum_{k: t^j_k < t^i_l} \exp(-\beta_{ij}(t^i_l - t^j_k))\right],
\end{multline}
where $t^{\max}\leq T$ is the largest jump time recorded over all  nodes.  
      
    \begin{remark}
    The function $l_i(\bx|\btheta_i)$ from \eqref{eq:log_lik} is convex in $\balpha_i$ and $\mu_i$.  Thus, the maximum likelihood estimate (MLE) can be computed  using convex optimization. The proof of convexity can be found, e.g. in \cite{ogata1981}.
    \end{remark}

\subsection{Hessian Matrix for Exp-MHPs}

In this section, we derive an explicit expression for the Hessian matrix $H(\btheta)$ of the negative log-likelihood $l(\bx|\btheta)$ based on formula \eqref{eq:log_lik}, see  \cite{Shlomovich2022}. Note that the Fisher information matrix $J$, required in the total message length criterion \eqref{eq:MML87_full}, is defined as the expected Hessian. In general, this expectation is difficult to compute and it is often replaced by the observed Fisher information. The observed Fisher information, in turn, is given by the Hessian, evaluated at an estimate $\hat{\btheta}$ of $\btheta$.

To obtain the Hessian, we first need to compute the gradient of the negative log-likelihood $l(\bx|\btheta)$.
The required first-order derivatives can be computed explicitly and are given by

\begin{align}
    \frac{\partial l(\bx|\btheta)}{\partial \mu_i} &= 
    t^{\max}- \sum_{l=1}^{n_i} \frac{1}{\mu_i + \sum_{k=1}^p \alpha_{ik}A_{ik}(t^i_l)}, \nonumber \\
    \frac{\partial l(\bx|\btheta)}{\partial \alpha_{ij}} &= \frac{1}{\beta_{ij}} \sum_{k=1}^{n_j} \left[ 1-\exp\left(-\beta_{ij}(
    t^{\max}- t^j_k) \right)\right] - \sum_{l=1}^{n_i} \frac{ A_{ij}(t_l^i) }{\mu_i + \sum_{k=1}^p \alpha_{ik}A_{ik}(t^i_l)},
\end{align}
where 
\begin{equation}
A_{ij}(t) = \sum_{k: t^j_k < t}\exp\left(-\beta_{ij}(t - t^j_k) \right), \quad \text{for} \ t \in [0,T].
\end{equation}
Intuitively, the term $A_{ij}(t)$ summarizes the ``weighted'' history of events on node $j$ up to time $t$: since the $\beta_{ij}$ are positive, ``new'' events will always have more importance than the ``old'' ones. 

Furthermore, the second order partial derivatives w.r.t. the parameters $\mu_i$ and $\alpha_{ij}$ are given by
\begin{equation}\label{eq:double_derivative}
\begin{aligned}
    \frac{\partial^2 l(\bx|\btheta)}{\partial \mu^2_i} &= \sum_{r=1}^{n_i} \frac{1}{\left(\mu_i + \sum_{k=1}^p \alpha_{ik}A_{ik}(t^i_r)\right)^2}, \\
     \frac{\partial^2 l(\bx|\btheta)}{\partial \mu_i \partial \alpha_{ij}} &= \sum_{r=1}^{n_i} \frac{A_{ij}(t^i_r)}{\left(\mu_i + \sum_{k=1}^p \alpha_{ik}A_{ik}(t^i_r)\right)^2}, \\ 
     \frac{\partial^2 l(\bx|\btheta)}{ \partial \alpha_{ij}\partial \alpha_{ij^\prime}} &= \sum_{r=1}^{n_i} \frac{A_{ij}(t^i_r) A_{ij^{\prime}}(t^i_r)}{\left(\mu_i + \sum_{k=1}^p \alpha_{ik}A_{ik}(t^i_r)\right)^2}.
\end{aligned}
\end{equation}

Note that all derivatives $\frac{\partial^2 l(\bx|\btheta)}{\partial \mu_i \partial \alpha_{i^\prime j}}, \frac{\partial^2 l(\bx|\btheta)}{ \partial \alpha_{ij}\partial \alpha_{i^\prime j^\prime}}$ for $i^\prime \neq i$ are equal to $0$.
Therefore, the Hessian of $l(\bx|\btheta)$ can be written as a block-diagonal matrix of the form 
\begin{equation}
    H(\btheta)=\begin{pmatrix}
     H_1(\btheta_1) & & \\
     & \ddots  & \\
     & & H_p(\btheta_p)
    \end{pmatrix},
\end{equation}
where, for each $i \in \{ 1,\ldots,p \}$,  
$H_i(\btheta_i)$ is given by the $(p+1)\times(p+1)$-dimensional matrix
\begin{equation}\label{eq:Hessian_i}
    H_i(\btheta_i)=\begin{pmatrix}
    \frac{\partial^2 l(\bx|\btheta)}{\partial \mu_i^2} & \frac{\partial^2 l(\bx|\btheta)}{\partial \mu_i \partial \alpha_{i1}} & \ldots & \frac{\partial^2 l(\bx|\btheta)}{\partial \mu_i \partial \alpha_{ip}} \\
    \frac{\partial^2 l(\bx|\btheta)}{ \partial \alpha_{i1} \partial \mu_i} & \frac{\partial^2 l(\bx|\btheta)}{\partial \alpha_{i1}^2} & \ldots & \frac{\partial^2 l(\bx|\btheta)}{\partial \alpha_{i1} \partial \alpha_{ip} } \\
    \vdots & \vdots & \ddots & \vdots \\
    \frac{\partial^2 l(\bx|\btheta)}{\partial \alpha_{ip} \partial \mu_i } & \frac{\partial^2 l(\bx|\btheta)}{ \partial \alpha_{ip} \partial \alpha_{i1} } & \ldots & \frac{\partial^2 l(\bx|\btheta)}{ \partial \alpha_{ip}^2 }
    \end{pmatrix},
\end{equation}
with entries as in \eqref{eq:double_derivative}.
Since the determinant of a block-diagonal matrix is equal to the product of the determinants of the diagonal blocks, we have that
\begin{equation}\label{eq:sum:Hessian}
    \log |H(\btheta)|= \sum_{i=1}^{p} \log |H_{i} (\btheta_i)|.
\end{equation}

Note that, in the case when the vector of intensities $\bmu$ is known, it is possible to use a more computationally efficient approximation of the Hessian, see \cite{wang}. We consider the intensities $\bmu$ to be unknown, thus we will rely on the analytical expression for the Hessian, with entries defined via \eqref{eq:double_derivative}.

\subsection{Choice of Priors for Exp-MHPs}\label{subsec:priors}

In this section, we define two possible prior distributions $\pi(\btheta)$ for the parameter vector $\btheta \in \bTheta$ \eqref{vectortheta2}, which is now considered to be a random quantity.

First, we assume that two parameter vectors $\btheta_i$ and $\btheta_j$ as in \eqref{vectortheta}, corresponding to different nodes $i \neq j$, are independent.
Therefore, the negative log-prior function can be expressed as
\begin{equation}\label{eq:sum:prior}
   -\log \pi (\btheta)=-\sum\limits_{i=1}^{p}  \log \pi_i(\btheta_i),
\end{equation}
where $\pi_i(\btheta_i)$ is a prior distribution for the parameter vector $\btheta_i \in \bTheta_i$ \eqref{vectortheta}, corresponding to the $i$-th node. 

Throughout, we further assume that $\mu_i$ and all entries $\alpha_{ij}$ of a parameter vector $\btheta_i$ \eqref{vectortheta} are independent and identically distributed (iid), yielding
\begin{equation}\label{eq:prior:nodei}
	\pi_i(\btheta_i)=\pi(\mu_i) \prod_{j=1}^{p}  \pi(\alpha_{ij}). 
\end{equation}

In the following, we consider two different prior distributions for the entries $\mu_i$ and $\alpha_{ij}$ of a parameter vector $\btheta_i \in (\mathbb{R}_0^+)^{p+1}$ as in \eqref{vectortheta}, which both allow to incorporate the prior knowledge that the parameters to be estimated are all non-negative: the uniform distribution $\textrm{U}[0,b]$, with $b>0$, and the exponential distribution $\textrm{Exp}(c)$, with $c>0$, having support $[0,\infty)$.

\paragraph{Uniform prior.}

Assuming that $\mu_i$ and the $\alpha_{ij}$ are iid as $U[0,b]$, $b>0$, the prior \eqref{eq:prior:nodei} for the $i$-th node becomes
\begin{equation}
\pi_i(\btheta_i)=\prod_{j=1}^{p+1} \frac{1}{b}=\frac{1}{b^{p+1}}.
\end{equation}
Thus, the negative log-prior for the $i$-th node is given by
\begin{equation}\label{eq:prior:uniform}
-\log \pi_i(\btheta_i)=(p+1)\log(b).
\end{equation}
Note that, to obtain a flat prior with a non-restrictive domain, one may consider a large value for the hyperparameter $b$ (see e.g. \cite{oliver99} for the use of uniform priors in the context of MML). 

\paragraph{Exponential prior.}

Assuming that $\mu_i$ and the $\alpha_{ij}$ are iid as $\textrm{Exp}(c)$, $c>0$, the prior \eqref{eq:prior:nodei} for the $i$-th node becomes
\begin{equation}\label{Eq_exp}
	\pi_i(\btheta_i)= c \exp(-c \mu_i)\prod_{j=1}^{p} c \exp(-c \alpha_{ij}) = c^{p+1} \exp \left( -c\mu_i -c\sum\limits_{j=1}^{p}\alpha_{ij} \right).  
\end{equation}
Thus, the negative log-prior for the $i$-th node is given by
\begin{equation}\label{eq:prior:exp}
	-\log \pi_i(\btheta_i)= c\mu_i + c\sum_{j=1}^{p} \alpha_{ij}-(p+1)\log(c). 
\end{equation}
In this case, to obtain a flat prior, one may choose a small value for the hyperparameter $c$.

\begin{remark}\label{rem2}
   The negative log-priors in \eqref{eq:prior:uniform} are constant in $\btheta_i$  and those in \eqref{eq:prior:exp} are linear in $\btheta_i$. Thus, they are convex in both cases. 
\end{remark}

\subsection{MML Criterion for Granger Causal Inference in Exp-MHPs}

From formulas \eqref{eq:sum:likelihood} and \eqref{eq:sum:prior}, it becomes 
evident that the optimization of the function $\log p(\bx|\btheta)+\log \pi(\btheta)$ w.r.t. $\btheta \in \bTheta$ can be done independently for each node $i$. Therefore, to perform causal inference in exp-MHPs, for each $i \in \{1,\ldots,p\}$, we introduce a structure set 
$\bGamma_i=\{0,1\}^p$, whose elements $\bgamma_i=(\gamma_{i1},\ldots,\gamma_{ip})^\top$ are $p$-dimensional vectors of zeros and ones with $k_i>0$ corresponding to the number of ones. It holds  then  that $\gamma_{ij}=1$ if and only if events in the $j$-th node Granger-cause events in the $i$-th node, and  $\gamma_{ij}=0$ if and only if $\alpha_{ij}=0$, i.e. there is no impact of node $j$ on node $i$. This means that causal discovery in exp-MHPs is equivalent to identifying the sparsity pattern in the influence vector $\balpha_i$, for each $i\in \{1,\ldots,p\}$.

According to the definition of Granger-causality in exp-MHPs, for a given structure $\bgamma_i \in  \bGamma_i=\{0,1\}^p$, the corresponding restricted parameter space $\bTheta_{\bgamma_i}$ contains parameter vectors representing $\mu_i$ and $\balpha_{i}$ such that $\alpha_{ij}$ is present in $\balpha_i$ if and only if $\gamma_{ij}=1$. Thus, for any $\bgamma_i$ containing $k_i$ non-zero entries, the vector $\balpha_i$ has $k_i$ non-zero entries to be estimated. Moreover, the baseline intensity $\mu_i$, which has no influence on causal discovery, has to be estimated as well. Hence, under a given structure $\bgamma_i$, a total of $k_i+1$ non-negative parameters are to be estimated and the restricted parameter space $\bTheta_{\bgamma_i}=(\mathbb{R}_0^+)^{k_i+1}$.

Recall that the formulas \eqref{eq:log_lik},
\eqref{eq:Hessian_i}, \eqref{eq:prior:uniform} and \eqref{eq:prior:exp} are formulated for $\btheta_i \in \bTheta_i=(\mathbb{R}_0^+)^{p+1}$ from \eqref{vectortheta}. For a given structure $\bgamma_i\in \bGamma_i= \{ 0,1 \}^p$ and parameter vector $\btheta_i \in \bTheta_{\bgamma_i}=(\mathbb{R}_0^+)^{k_i+1}$ with $\balpha_i$ having $k_i\leq p$ entries, the log-likelihood $\log p_{\bgamma_i}(\bx|\btheta_i)$, 
the Hessian $H_{\bgamma_i}(\btheta_i)$ and log-priors $\log \pi_{\bgamma_i}(\btheta_i)$ are obtained from the aforementioned formulas, replacing $p$ by $k_i$  and adjusting all indices properly. 

Finally, for each node $i \in \{1,\ldots,p\}$, we can now propose the following refined message length criterion for causal inference in exp-MHPs:
\begin{equation}\label{eq:MML87_full_expMHP_k}
\begin{aligned}
I(\bx;\btheta_i;\bgamma_i) 
= & - \log p_{\bgamma_i}({\bx}|\btheta_i)  - \log \pi_{\bgamma_i}(\btheta_i) + \frac{1}{2} \log|H_{\bgamma_i}(\hat{\btheta}_i)| \\
& -\frac{k_i}{2} \log(2\pi) + \frac{1}{2} \log(k_i\pi) +  \psi(1) \\
& + \log \dbinom{p}{k_i} + \log(p +1),
\end{aligned}
\end{equation}
where, for a given structure $\bgamma_i\in \bGamma_i=\{0,1\}^p$, the Hessian matrix is evaluated at the estimate $\hat{\btheta}_i$ given by
\begin{equation*}
    \hat{\btheta}_i =  \textrm{argmin}_{\btheta_i \in \Theta_{\gamma_i}} \left( -\log p_{\bgamma_i}(\bx | \btheta_i)-\log \pi_{\bgamma_i}(\btheta_i) \right).
\end{equation*}

\begin{remark}
i) Note that under the uniform prior \eqref{eq:prior:uniform} the estimate $\hat{\btheta}_i$ coincides with the MLE. ii) One may either remove the $p$-dimensional zero vector from the structure set $\bGamma_i$ or allow for $k_i=0$ (node $i$ does not receive any input connections) by setting the term $k_i(\log \kappa_{k_i}+1)/2$ (cf. formula \eqref{eq:approx}) to zero in that case.  Criterion \eqref{eq:MML87_full_expMHP_k} for $k_i=0$ then results into the form, where the second line is not present.
\end{remark}

\section{Algorithm MMLH for Granger Causal Inference in Exp-MHPs and its  Complexity}\label{sec:algorithm}

The form of the MML criterion (\ref{eq:MML87_full_expMHP_k}) leads to the following algorithm, denoted as MMLH, which we propose for causal inference in exp-MHPs.

\begin{algorithm}[H]
	\caption{MMLH: Causal inference in exp-MHPs by MML
		\ \\ \textbf{Input:} Dimension $p$, data $\bx$ \ \\
		\textbf{Output:} Estimate $\hat{\bgamma}=[\hat{\bgamma}_1^\top,\ldots,\hat{\bgamma}_p^\top]^\top \in \bGamma:=\bGamma_1\times \ldots \times \bGamma_p$
	}\label{alg:MML:expMHP}
	\begin{algorithmic}[1]
		\For{each $i \in \{1, \ldots, p$ \}}
	\For{each $\bgamma_i \in \bGamma_i=\{ 0,1 \}^{p}$}
		\State $\hat{\btheta}_i \longleftarrow \textrm{argmin}_{\btheta_i \in \bTheta_{\gamma_i}} \left( -\log p_{\bgamma_i}(\bx | \btheta_i)-\log \pi_{\bgamma_i}(\btheta_i) \right)$
		\State $\hat{c}_{\bgamma_i} \longleftarrow I(\bx;\hat{\btheta}_i;\bgamma_i)$ \ \eqref{eq:MML87_full_expMHP_k}
		\EndFor
	\State  $\hat{\bgamma}_i \longleftarrow \textrm{argmin}_{\bgamma_i \in \bGamma_i} \ \hat{c}_{\bgamma_i} $
	\EndFor
	\State \textbf{return} $\hat{\bgamma}=[\hat{\bgamma}_1^\top,\ldots,\hat{\bgamma}_p^\top]^\top$
	\end{algorithmic}\label{alg:MMLH}
\end{algorithm}

\begin{remark}
    Recall that the decay constants $\beta_{ij}$ are assumed  to be known (see Section \ref{section:model}). 
    However, adding more unknown parameters would be possible and would require small modifications in Algorithm \ref{alg:MMLH}.
    The estimation procedure in lines 3 and 4  would have to be adjusted accordingly, by extending the parameter space and setting priors for the $\beta_{ij}$. Moreover, the Hessian matrix would include double derivatives with respect to $\beta_{ij}$, which are available as closed-form expressions in the literature (see \cite{ozaki79}).
\end{remark}

We now address the computational complexity of the proposed method. Algorithm~1 consists of $p$ optimizations, each requiring $2^{p}$  evaluations of the MML criterion  \eqref{eq:MML87_full_expMHP_k}. Each such evaluation relies on a parameter learning procedure for the observed data~$\bx$ (line 3 of Algorithm~\ref{alg:MMLH}) and a function evaluation (line 4 of Algorithm~\ref{alg:MMLH}), which includes a computation of the corresponding Hessian matrix and its determinant.

The computational complexity of a parameter learning procedure (line 3) depends on the number of parameters to be estimated (for a given structure $\bgamma_i \in \bGamma_i=\{0,1\}^p$ with $k_i$ non-zero entries, $k_i+1$ parameters have to be estimated and $k_i\leq p$) and the size of $\bx$, i.e. the number of observed events (which in turn depends on $T$). In our R-implementation, we apply the Nelder-Mead search method (NM), which relies on a user-defined termination.  There is no convergence theory  providing
an estimate for the number of iterations required to satisfy a reasonable accuracy
constraint given in the termination test, see \cite{singer1999complexity}. Thus,   to the best of our knowledge, an  upper bound on the complexity of the NM method with a fixed termination rule is not~available~in~the~literature. 

The complexity of computing the determinant of the symmetric Hessian matrix (line~4) also depends on the ``size'' of $\bgamma_i$, the dimension of $H_{\bgamma_i}$ being $(k_i+1)\times(k_i+1)$. In our R-implementation, we use a LU decomposition procedure, typically having a complexity of order $(k_i+1)^3$ (note that \cite{aho1974} showed, however, that the exponent $3$ may be reduced to $2.373$ via a  fast matrix multiplication method).

In scenarios where the expert knowledge suggests an upper bound on the number of causes for each node $i \in \{1,\ldots,p\}$, the computational complexity of MMLH can be reduced. Concretely, assume that for each node $i$ the structure space $\bGamma_i=\{ 0,1 \}^p$ only contains $p$-dimensional binary vectors with at most $m<p$ non-zero entries. In this case, for the cardinality of $\bGamma_i$ it holds that
\vspace{-0.2cm}
\begin{equation*}
    | \bGamma_i | = \sum_{k=0}^m \binom{p}{k} < m\binom{p}{m} = {O}(p^m).
    \vspace{-0.2cm}
\end{equation*}
Therefore, the number of required evaluations of the MML criterion \eqref{eq:MML87_full_expMHP_k} reduces from $p 2^p$ to less than $p p^m=p^{m+1}$, which may be beneficial especially for large dimensions $p$ and upper bounds $m \ll p$. Moreover, since now $k_i \leq m < p$, also the computational cost of each parameter learning procedure (line 3) and determinant computation (line 4) reduces. Note that the upper bound $m$ may be given as 
input parameter to Algorithm~\ref{alg:MMLH} and the corresponding reduction of the structure set 
(from $p$-dim. binary vectors with at most $p$ non-zero entries to those with at most $m<p$ non-zero entries) is straight-forward~to~implement.

Note also that, in the general framework of a non-reduced structure space, the related state-of-the-art MDLH method (\cite{jalaldoust}) also requires to solve $p$ optimization problems, each relying on $2^p$ evaluations of their MDL function. Such an evaluation also contains a parameter estimation procedure for the observed data~$\bx$. In addition, while this function does not include Hessian matrix and determinant computations, it requires a large number $N$ of Monte Carlo simulations for additional parameter learning (integral estimation). This results in a total of $(N+1) p 2^{p}$ parameter learning procedures required~by~MDLH.

\vspace{-0.15cm}
\begin{remark}\label{rem:par}
    In order to optimize the performance of the MMLH method, especially in high-dimensional setups, it is recommended to carry out the $p$ independent optimizations (lines 2--6 of Algorithm 1) in parallel. Additionally, note that when the number of nodes is large, it is possible to replace the exhaustive search algorithm by a genetic algorithm, which can be parallelized as well, see e.g. \cite{hlavackova2020}.
    \vspace{-0.2cm}
\end{remark}

\section{Numerical Experiments}\label{section:numerical_experiments}

In this section, we illustrate the performance of the proposed MMLH algorithm on both simulated synthetic data (with known ground-truth connectivity matrices) and real G7 sovereign bond data.

In all our experiments, we consider both MMLH with uniform prior \eqref{eq:prior:uniform} (denoted by MMLH-u) and with exponential prior \eqref{eq:prior:exp} (denoted by MMLH-e). These proposed procedures are compared with the related state-of-the-art MDL-based method for causal inference in exp-MHPs (denoted by MDLH) introduced in \cite{jalaldoust}. As a representative of the lasso-based procedures we consider the method ADM4 from \cite{zhou2013learningb}. Note that this method has been also considered in the experiments reported in  \cite{jalaldoust} and shown to outperform e.g. the method NPHC from \cite{achab2017}. Moreover, we consider two classical related model selection methods, namely the one obtained via the BIC and the one obtained via the AIC (i.e. Algorithm \ref{alg:MMLH} where the criterion~ \eqref{eq:MML87_full_expMHP_k} is replaced by the BIC and AIC, respectively). Further, we investigate Algorithm~\ref{alg:MMLH} where the criterion~\eqref{eq:MML87_full_expMHP_k} is reduced to the first term only (the negative log-likelihood). This method is denoted by MLE-ms, where ``ms'' stands for model selection. We also consider standard maximum likelihood estimation, where node $j$ is assumed to cause node $i$ if and only if the estimated $\alpha_{ij}$-value is larger than a pre-set threshold (here $0.1$). This method is denoted by MLE-thr. The following results for MDLH and ADM4 are based on the code provided by \cite{jalaldoust}, the respective algorithms of the other methods are newly implemented. 

\vspace{-0.2cm}
\subsection{Experiments with Synthetic Data}
\vspace{-0.1cm}

In our synthetic experiments, we choose different setups inspired by those reported in \cite{jalaldoust}. In particular, we investigate exp-MHPs of dimension $p=7$, $10$ and $20$, respectively, and focus on short time horizons $T$, i.e. $T\leq 100p$. Moreover, we use the F1 score to evaluate the accuracy of our inferred connectivity matrices (in comparison to the respective ground truth). 
The $F1$ score is defined as the harmonic mean of the precision and recall measures: $$\text{F1 score}= \frac{2 \cdot \mbox{precision} \cdot \mbox{recall}}{\mbox{ precision} + \mbox {recall }},
\vspace{-0.2cm}
$$ 
where
\vspace{-0.2cm}
\begin{align*}
    \mbox{precision}&= \frac{\text{number of correctly predicted edges (ones)}}{\text{total number of predicted edges (ones)}},\\
    \mbox{recall}&= \frac{\text{number of correctly predicted edges (ones)}}{\text{number of edges (ones) present in the ground truth}}.
\end{align*} 

In all considered settings, the experiments are repeated $N$ (here, $N=100$) times, yielding $N$ estimates $\hat\bgamma^1,\ldots, \hat\bgamma^N$ from Algorithm~\ref{alg:MMLH}. An average F1 score over the $N$ trials, along with the corresponding standard deviation (put in parenthesis), is reported. The reported computing times are also averaged over trials and correspond to an implementation of Algorithm \ref{alg:MMLH}, which is parallelized at the level of nodes (see Remark \ref{rem:par}). The code was run on $p$ parallel cores of a HPC architecture located at the University of Klagenfurt (AMD EPYC 7532, 2.4 GHz, 32-core processor).

We focus on sparse connectivity graphs (two different settings) and investigate also the mid-dense setting considered in \cite{jalaldoust}.

\paragraph{Sparse settings}

We consider two different sparse settings. In the first one, the causal structure corresponds to a unidirectional (cascade) coupling structure with self-excitation in the first component. This means that all entries $\alpha_{ij}$ of the influence matrix $\balpha$ are zero, except for $\alpha_{11}$ and those in the lower diagonal (i.e., the entries $\alpha_{(i+1),i}$, $i=1,\ldots,p-1$). In the second setting, each node is influenced either by itself or by one of the other components (single input structure).  This means that the influence matrix~$\balpha$ has exactly one non-zero entry per row, which (in contrast to the cascade setting) is randomly placed. 

Both settings have $p$ connections (out of a total of $p^2$ possible connections), corresponding to $14.3\%$, $10\%$ and $5 \%$ of edges for $p=7$, $10$, and $20$, respectively. For $p=10$ and $p=20$, we assume to have some prior expert knowledge on the maximum number of causes per node and reduce the model search to structure sets $\bGamma_i=\{0,1\}^p$, $i \in \{1,\ldots,p\}$, which contain binary vectors with at most $m=5$ and $m=3$ non-zero entries, respectively. Moreover, in both settings we set all non-zero $\alpha_{ij}$-parameters to $0.55$, and consider $\mu_i=0.5$ and $\beta_{ij}=1$. 
Furthermore, we set the parameter of the uniform prior \eqref{eq:prior:uniform} and exponential prior~\eqref{eq:prior:exp} to $b=10^5$ and $c=10^{-5}$, respectively, choosing thus very flat and little-informative~prior~distributions.  

The results for the two sparse settings are reported in Table \ref{table:cascade} and Table \ref{table:oneInput}, respectively, and also compared to those obtained by randomly assigning one connection per row in the desired connectivity matrix, a procedure denoted by RAND.  
We observe that both variants of the proposed algorithm MMLH-u and MMLH-e yield F1 scores higher than those of the other methods 
and that there is no tangible difference between these two variants. Moreover, the results obtained with MMLH are comparable to those obtained with BIC. This may be explained by the fact that a large value of $b$ (resp. small value of $c$) leads to a stronger penalty of structures $\bgamma_i \in \bGamma_i = \{0,1\}^p$ with large number of non-zero entries $k_i$, as it happens in BIC. Another observation is the poor performance of the classical maximum likelihood approach MLE-thr, especially for large dimensions $p$. This may result from the fact that, when $p$ is large, a lot of parameters have to be estimated simultaneously under this method, while the model selection approach reduces the amount of parameters to be estimated, taking different structures into account. Moreover, especially in sparse settings, the limited time horizon may not allow for enough observations to ensure the convergence of the maximum likelihood estimator to the true parameter vector in practice.

\begin{figure}
    \centering
        \includegraphics[width = 0.87\textwidth]{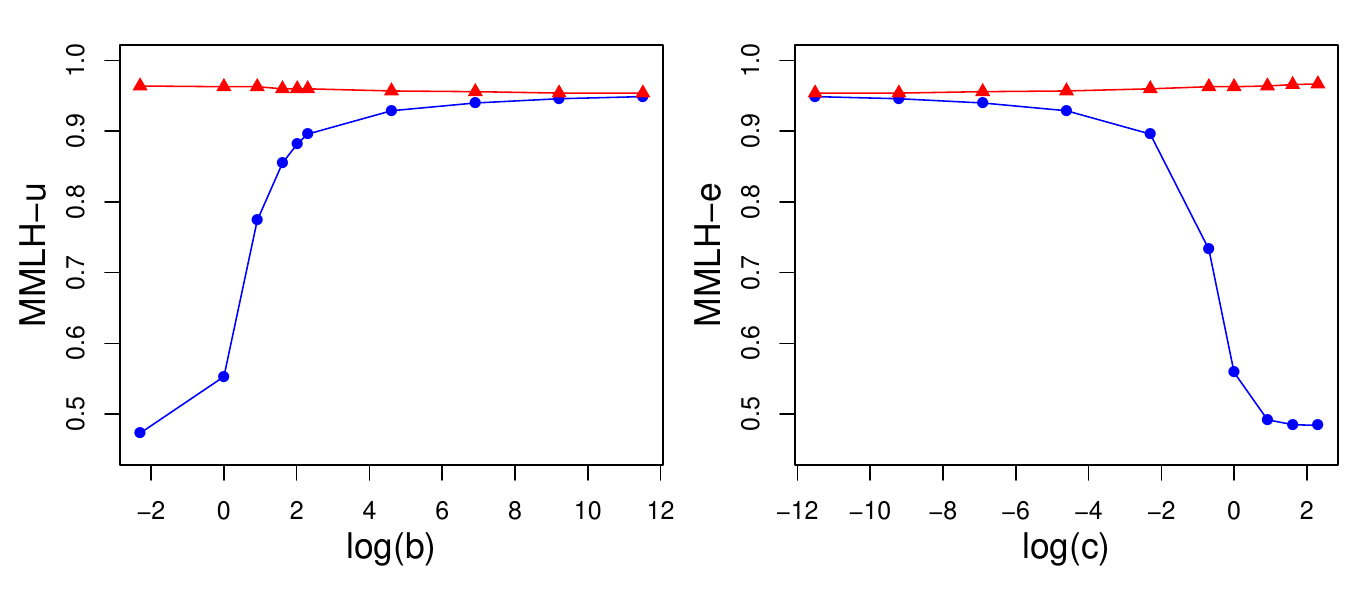}
    \caption{Cascade structure for $p=7$ and $T=200$. F1 score (blue lines) and TP-score (red lines) as functions of the uniform prior parameter $b$ (left panel) and exponential prior parameter $c$ (right panel). The x-axes are reported in log-scale.} 
    \label{fig:F1score_bc}
\end{figure}

The impact of the choice of $b$ (resp. $c$) for MMLH-u (resp. MMLH-e) on the F1 score is illustrated in Figure \ref{fig:F1score_bc} (blue lines), where we focus on the cascade scenario with $p=7$ and $T=200$. Decreasing $b$ (resp. increasing $c$) leads to a decrease in the F1 accuracy. However, we observe that the ``true positive'' (TP) score (red lines) is not strongly influenced by the choice of $b$ (resp. $c$). This means that when decreasing $b$ (resp. $c$), MMLH still identifies the correct connections with a high precision, but also proposes connections which are not present in the underlying ground truth graph. Similar observations can be made for the second sparse setting (figures not shown).

\begin{table}[H]
\tabcolsep=0.055cm
    \hspace{-0.5cm}{\footnotesize{\begin{tabular}{|r|ccc|ccc|c|}
    \hline
    & \multicolumn{7}{c|}{F1 score 
} \\
    \hline
      $p=$ & \multicolumn{3}{c|}{7} &  \multicolumn{3}{c|}{10 (\mbox{red.} $\bGamma_i$, m=5) } & \multicolumn{1}{c|}{20 (\mbox{red.} $\bGamma_i$, m=3)}  \\
        \hline 
        $T =$ & $200$ & $400$& $700$  & $200$ & $400$ & $700$ & $200$ \\
         \hline
        Runtime & 12.7 s & 40.4 s & 116.4 s & 78.0 s & 277.4 s & 821.3 s & 521.5 s \\
         \hline
        MMLH-u & \textbf{0.948} {(0.089)} & \textbf{0.979} {(0.053)} & \textbf{0.985} {(0.046)} &     \textbf{0.953} {(0.072)} & \textbf{0.968} {(0.062)} & \textbf{0.982} {(0.039)} & \textbf{0.933} {(0.067)} \\ 
        MMLH-e & \textbf{0.948} {(0.089)} &  \textbf{0.979} {(0.053)}& \textbf{0.985} {(0.046)} &    \textbf{0.953} {(0.072)} & \textbf{0.968} {(0.062)} & \textbf{0.982} {(0.039)} &  \textbf{0.933} {(0.067)} \\ 
        MLE-ms & 0.572 {(0.048)}  & 0.603 {(0.038)} & 0.623 {(0.035)} & 0.540 {(0.034)} & 0.577 {(0.038)} & 0.602 {(0.032)} & 0.505 {(0.022)} \\
        MLE-thr & 0.410 {(0.082)} &  0.416 {(0.069)} & 0.413 {(0.078)} &  0.231 {(0.052)} & 0.240 {(0.052)} & 0.236 {(0.051)} & 0.099 {(0.017)} \\
        BIC & 0.943 {(0.090)} &  0.977 {(0.055)} & 0.983 {(0.046)} &  0.943 {(0.080)} & 0.964 {(0.063)} & 0.977 {(0.042)} & 0.927 {(0.068)} \\
        {AIC} & {0.850} {(0.094)} &  {0.862} {(0.079)} & {0.896} {(0.076)} &  {0.804} {(0.086)} & {0.833} {(0.074)} & {0.860} {(0.064)}  & {0.719} {(0.047)} \\
RAND &  0.130 {(0.117)} &  0.159 {(0.150)} & 0.159 {(0.131)} &  0.102  {(0.096)} & 0.095 {(0.097)} & 0.084 {(0.093)} & 0.045 {(0.045)} \\
         ADM4 & 0.773   {(0.063)} &  0.782  {(0.055)} &  0.807 {(0.049)} & 0.733 {(0.039)} & 0.759 {(0.034)} & 0.784 {(0.050)} & 0.694 {(0.051)} \\
         MDLH & 0.566 {(0.062)} & 0.533 {(0.062)} & 0.556 {(0.061)} & 0.431 {(0.056)} & 0.421 {(0.041)} & 0.429 {(0.060)} & 0.398 {(0.046)} \\
        \hline
    \end{tabular}} }
     \caption{Sparse setting 1: Cascade structure. The values for the uniform and exponential priors are $b=10^5$ and $c=10^{-5}$, respectively. Moreover, red. $\bGamma_i$ denotes a reduced structure set.} \label{table:cascade}
\end{table}

\begin{table}[H]
   \tabcolsep=0.055cm
    \hspace{-0.5cm}{\footnotesize{\begin{tabular}{|r|ccc|ccc|c|}
    \hline
    & \multicolumn{7}{c|}{F1 score 
} \\
    \hline
      $p=$ & \multicolumn{3}{c|}{7} &  \multicolumn{3}{c|}{10 (\mbox{red.} $\bGamma_i$, m=5) } & \multicolumn{1}{c|}{20 (\mbox{red.} $\bGamma_i$, m=3)}  \\
        \hline 
        $T =$ & $200$ & $400$& $700$  & $200$ & $400$ & $700$ & $200$ \\
          \hline
        {Runtime} & {13.0 s} & {41.2 s} & {115.1 s}  & {79.5 s} & {278.9 s} & {836.1 s} & {522.2 s} \\
         \hline
        MMLH-u & \textbf{0.956} {(0.074)} & \textbf{0.967} {(0.068)} & \textbf{0.978} {(0.051)} &     \textbf{0.944} {(0.084)} & \textbf{0.960} {(0.070)} & \textbf{0.958} {(0.066)} & \textbf{0.929} {(0.072)} \\ 
        MMLH-e & \textbf{0.956} {(0.074)} & \textbf{0.967} {(0.068)} & \textbf{0.978} {(0.051)} &     \textbf{0.944} {(0.084)} & \textbf{0.960} {(0.070)} & \textbf{0.958} {(0.066)} & \textbf{0.929} {(0.072)} \\ 
        MLE-ms & 0.571 {(0.043)} & 0.599 {(0.039)} & 0.617 {(0.032)} &  0.536 {(0.037)} & 0.569 {(0.040)} & 0.592 {(0.035)} & 0.502 {(0.025)}\\
        MLE-thr & 0.414  {(0.061)} &  0.404 {(0.081)} & 0.401 {(0.077)} &  0.243 {(0.052)} & 0.237 {(0.054)} & 0.234 {(0.059)} & 0.096 {(0.021)} \\
        BIC & 0.954 {(0.075)} &  0.961 {(0.072)} & 0.975 {(0.056)} &  0.936 {(0.085)} & 0.954 {(0.073)} & 0.955 {(0.068)} & 0.922 {(0.073)} \\
        {AIC} & {0.849} {(0.082)} &  {0.863} {(0.089)} & {0.875} {(0.080)} &  {0.794} {(0.075)} & {0.820} {(0.081)} & {0.837} {(0.070))} & {0.710 (0.053)} \\
        RAND &  0.127 {(0.128)}  &  0.154 {(0.149)} & 0.144 {(0.131)} &   0.093 {(0.082)} & 0.091 {(0.090)} & 0.104 {(0.096)}  & 0.052 {(0.047)} \\
         ADM4 & 0.471 {(0.068)}&   0.528 {(0.043)}  & 0.555 {(0.059)} & 0.519 {(0.073)} & 0.516 {(0.063)} & 0.529 {(0.067)} & 0.353 {(0.042)} \\
          MDLH & 0.768 {(0.066)} & 0.828  {(0.076)} & 0.872 {(0.060)} & 0.742 {(0.070)} & 0.780 {(0.071)} & 0.835 {(0.047)} & 0.686 {(0.041)} \\
        \hline
    \end{tabular}}}
     \caption{Sparse setting 2: Single input structure. The values for the uniform and exponential priors are $b=10^5$ and $c=10^{-5}$, respectively. Moreover, red. $\bGamma_i$ denotes a reduced structure set.}\label{table:oneInput}
\end{table}

\vspace{-0.5cm}
\paragraph{Mid-dense setting}

Now, we investigate a default scenario considered in \cite{jalaldoust}. In this setting, all diagonal entries of the influence matrix $\balpha$ are non-zero, i.e. all nodes are self-excitatory. All non-diagonal entries of the adjacency matrix of the underlying connectivity graph are randomly drawn from a Bernoulli distribution with success probability 0.3. In the case of a success (one), the corresponding $\alpha_{ij}$ is drawn from $U([0.1,0.2])$ (and so are the $\alpha_{ii}$). Moreover, each entry $\mu_i$ of the baseline vector $\bmu$ is drawn from $U([0.5,1.0])$ and all $\beta_{ij}$ are again set to $1$. Here, we further set the prior parameters $b$ and $c$ to $4$ and $0.3$, respectively, reducing the penalty strength on structures $\bgamma_i$ with a larger number of non-zero entries. This is motivated by the fact that this scenario may be considered as a mid-dense setting, since the ground truth connectivity matrices contain on average $p+0.3p(p-1)$ connections. In the case of $p=7$, this corresponds to an average of $40\%$~of~edges.

The results are reported in Table~\ref{table:middense} for different values of $T$. We observe that the method with the highest F1-accuracy is MDLH. Moreover, on shorter time horizons $T$, MMLH is also outperformed by ADM4 and MLE-thr.  For $T\geq 1000$, the two rivals are MDLH and AIC, which give a score close to $0.95$ and $0.92$, respectively (MMLH, in comparison, is approaching $0.9$). For the chosen prior parameters, we observe a slightly better performance for MMLH-u than for MMLH-e, except for the cases $T=200$ and $T=400$. 

\begin{table}
\tabcolsep=0.13cm
    \centering
    \footnotesize{\begin{tabular}{|r|cccccc|}
    \hline
    & \multicolumn{6}{c|}{F1 score 
} \\
\hline
        $p=$ & \multicolumn{6}{c|}{7}     \\
        \hline 
        $T =$ & $200$ & $400$& $700$ &  $1000$ & $1200$& $1400$   \\
         \hline
        {Runtime} & {54.3 s} & {149.4 s} & {472.4 s}  & {882.2 s} & {1818.2 s} & {2354.5 s} \\
         \hline
        MMLH-u & 0.567 {(0.106)} & 0.692 {(0.084)} & 0.787 {(0.070)} & 0.833 {(0.059)}  & 0.857 {(0.054)} & 0.872 {(0.049)} \\
        MMLH-e & 0.579 {(0.096)} & 0.692 {(0.085)} & 0.772 {(0.068)} & 0.812 {(0.060)}  & 0.841 {(0.059)} & 0.847 {(0.050)}  \\
        MLE-ms & 0.597 {(0.080)} & 0.670 {(0.087)} & 0.797 {(0.081)} &  0.767 {(0.065)} & 0.795 {(0.070)} & 0.804  {(0.052)}  \\
          MLE-thr & 0.623  {(0.096)} & 0.729 {(0.097)} & 0.786 {(0.096)} &  0.824 {(0.075)} & 0.836 {(0.079)} & 0.863  {(0.065)}  \\
           BIC & 0.399 {(0.080)} & 0.471 {(0.086)} & 0.537 {(0.075)} &  0.613 {(0.071)} & 0.645 {(0.074)} & 0.707 {(0.070)}  \\
           {AIC} & {0.490} {(0.099)} & {0.660} {(0.103)} & {0.802} {(0.081)} &  {0.877} {(0.065)} & {0.901} {(0.059)} & {0.920} {(0.040)}  \\
             ADM4 & 0.695 {(0.072)} & 0.748 {(0.066)} & 0.761 {(0.067)} &  0.786 {(0.059)}  & 0.784 {(0.056)} & 0.786  {(0.063)} \\
             MDLH & \textbf{0.767} {(0.062)}  & \textbf{0.841} {(0.058)} & \textbf{0.900} 
             {(0.052)} &  \textbf{0.927} {(0.041)} & \textbf{0.936} {(0.046)} & \textbf{0.942} {(0.035)}  \\
        \hline
    \end{tabular}}
    \caption{Mid-dense setting 3: Bernoulli random structure. The values for the uniform and exponential priors are $b=4$ and $c=0.3$, respectively. 
    }\label{table:middense}
\end{table}

\begin{figure}
    \centering
       \includegraphics[width = 0.87\textwidth]{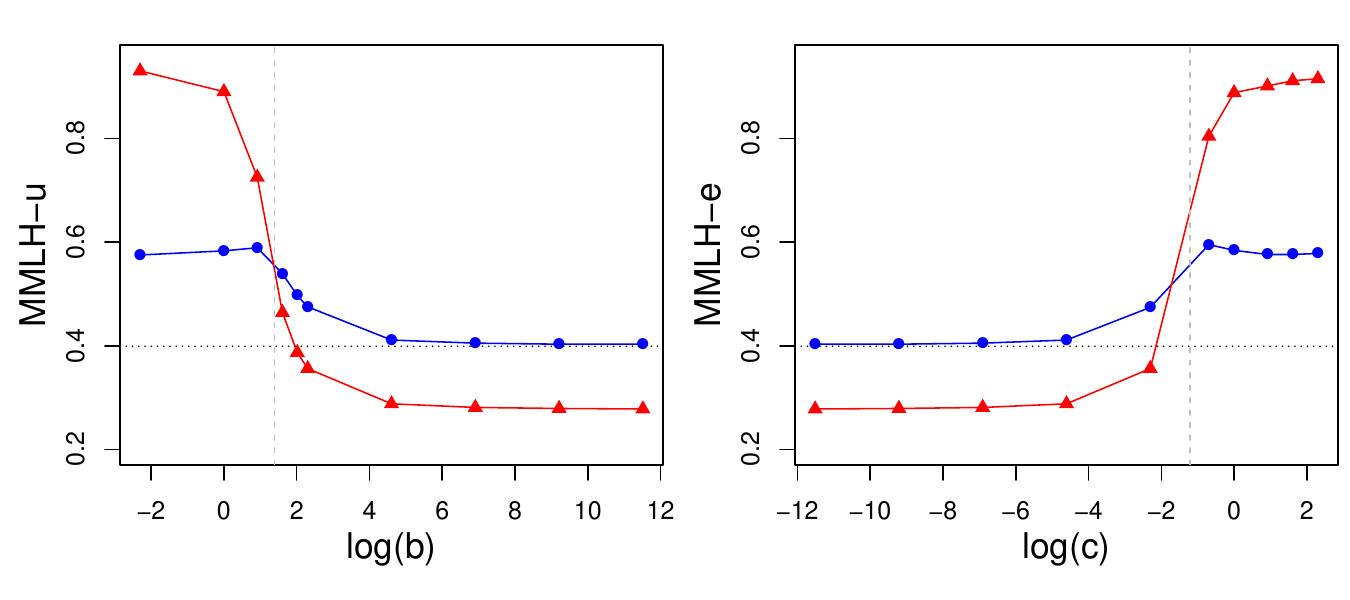}
    \caption{Bernoulli random structure for $p=7$ and $T=200$. F1 score (blue lines) and TP score (red lines) as functions of the uniform prior parameter $b$ (left panel) and exponential prior parameter $c$ (right panel). The vertical grey dashed lines indicate the investigated values $b=4$ and $c=0.3$, and the horizontal black dotted lines correspond to the F1 score for BIC. The x-axes are reported in log-scale.} 
    \label{fig:F1score_bc_midDense}
\end{figure}

In Figure \ref{fig:F1score_bc_midDense}, we report again the impact of $b$ (left panel) and $c$ (right panel) on the F1 score (blue lines) and TP score (red lines) for the case $T=200$. The previously considered values $b=4$ and $c=0.3$ are marked as vertical grey dashed lines. Remarkably, while in the sparse scenarios BIC almost reached the performance of MMLH, both MMLH-u and MMLH-e outperform BIC in this mid-dense setting for all considered values of $b$ and $c$ (though only slightly for large $b$ and small $c$).

\paragraph{Impact of the choice of the decay constants}

Now we study how the choice of the decay constants $\beta_{ij}$ influences the performance of the considered methods. We focus on the first sparse setting (cascade structure) and note that similar observations are made also for the other previously investigated scenarios. In particular, we consider the left column of Table \ref{table:cascade} (where $T=200$ and $\beta_{ij}=1$) and analyse how the results reported there change for other choices of $T$ and $\beta_{ij}$, see Table \ref{table:beta}. Note that, the larger the decay constants $\beta_{ij}$, the fewer event occurrence times are observed on a fixed time horizon $T$. Thus, in the first and third columns of Table \ref{table:beta} fewer data points are observed on average, while in the second and fourth column of Table \ref{table:beta} the average number of observed data points is approximately the same as in the left column of Table \ref{table:cascade}.

It is observed that all methods perform worse for larger values of the decay constants $\beta_{ij}$, except for MDLH whose performance seems not to be strongly influenced by the choice of the decay constants. However, there are no changes in the ranking of the considered methods (except for MDLH slightly outperforming ADM4 when $\beta_{ij}=1.5$). Moreover, we find that larger values of the time horizon $T$, to adjust for the decrease in the average number of observed data points caused by larger values of the decay constants $\beta_{ij}$, do not fully compensate for the loss in performance. Note that these relationships can be also observed for other values of the $\beta_{ij}$.

\begin{table}[H]
    \centering
    \normalsize{\begin{tabular}{|r|cccc|}
    \hline
    & \multicolumn{4}{c|}{   F1 score 
} \\
\hline
         $p= 7$, $T $ & 200 & 280  &  200 & 324 \\
         \hline
        $\beta_{ij} $ & 1.5 & 1.5  &  2 & 2 \\
         \hline
        MMLH-u & \textbf{0.837} (0.155) & \textbf{0.857} (0.132) & \textbf{0.715} (0.158) & \textbf{0.798} (0.148) \\
        MMLH-e & \textbf{0.837} (0.155) & \textbf{0.857} (0.132) & \textbf{0.715} (0.158) & \textbf{0.798} (0.148) \\
        MLE-ms & 0.542 (0.068) & 0.552 (0.063) & 0.484 (0.071) & 0.529 (0.064) \\
          MLE-thr & 0.342 (0.048) & 0.347 (0.052) & 0.352 (0.051) & 0.359 (0.044) \\
           BIC & 0.834 (0.149) & 0.852 (0.133) & 0.709 (0.155) & 0.790 (0.145) \\
          AIC & 0.762 (0.134) & 0.769 (0.112) & 0.664 (0.137) & 0.723 (0.132) \\
             ADM4 & 0.597 (0.060)  & 0.612 (0.050) & 0.620 (0.051) &  0.642 (0.049)  \\
             MDLH & 0.617 (0.073)  & 0.615 (0.080) & 0.608 (0.082) &  0.624 (0.074) \\
        \hline
    \end{tabular}}
    \caption{Sparse setting 1: Cascade structure. Different values for the decay constants $\beta_{ij}$ and time horizon $T$ are considered. The values for the uniform and exponential priors are $b=10^5$ and $c=10^{-5}$, respectively.}\label{table:beta}
\end{table}

\subsection{Experiments with Real-World Data} 

The goal of this subsection is to illustrate how the proposed MMLH approach performs on real-world data. In particular, we consider 10-year (2003-2014) sovereign bond yield volatilities of seven large economies called the Group of Seven (G7), being composed of the US, Canada, Germany, France, Japan, UK and Italy. This dataset has been investigated in \cite{demirer2018estimating} and also analysed in \cite{jalaldoust}. It is publicly available at: \url{http://qed.econ.queensu.ca/jae/2018-v33.1/demirer-et-al/}

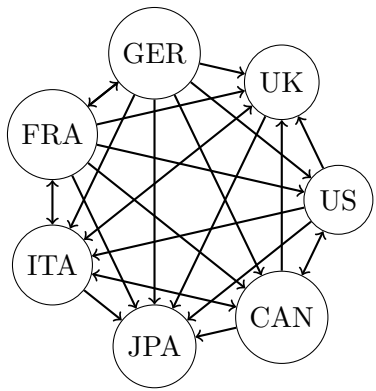
\begin{figure}[ht]
\centering
\begin{tikzpicture}
\def \n {7}
\def \radius {2.0cm}
\def \margin {8} 
  \node[draw, circle] at ({360/7*0}:\radius ) (US) {US};
\node[draw, circle] at ({360/7*1}:\radius ) (UK) {UK};
\node[draw, circle] at ({360/7*2}:\radius ) (GER) {GER};
\node[draw, circle] at ({360/7*3}:\radius ) (FRA) {FRA};
\node[draw, circle] at ({360/7*4}:\radius ) (ITA) {ITA};
\node[draw, circle] at ({360/7*5}:\radius ) (JPA) {JPA};
\node[draw, circle] at ({360/7*6}:\radius ) (CAN) {CAN};
\draw[->, thick] (FRA) -- (GER);
\draw[->, thick] (GER) -- (FRA);
\draw[->, thick] (GER) -- (UK);
\draw[->, thick] (GER) -- (US);
\draw[->, thick] (GER) -- (CAN);
\draw[->, thick] (GER) -- (JPA);
\draw[->, thick] (GER) -- (ITA);
\draw[<->, thick] (FRA) -- (ITA);
\draw[->, thick] (FRA) -- (UK);
\draw[->, thick] (FRA) -- (US);
\draw[->, thick] (FRA) -- (CAN);
\draw[->, thick] (FRA) -- (JPA);
\draw[<->, thick] (ITA) -- (UK);
\draw[->, thick] (ITA) -- (JPA);
\draw[<->, thick] (CAN) -- (ITA);
\draw[->, thick] (CAN) -- (JPA);
\draw[->, thick] (CAN) -- (UK);
\draw[<->, thick] (US) -- (CAN);
\draw[->, thick] (US) -- (JPA);
\draw[->, thick] (US) -- (UK);
\draw[->, thick] (US) -- (ITA);
\draw[->, thick] (UK) -- (JPA);
\end{tikzpicture}
\caption{Connectivity graph derived from \cite{umar2022network} (an edge is drawn if it appears in at least one of the three graphs shown in their Figure 1).}\label{fig:umar} 
\end{figure}

As this dataset corresponds to time series data, we pre-process it to identify shocks (i.e. events able to be reproduced by a point process) in the return volatilities. In particular, following \cite{jalaldoust}, for each country we roll a one-year window over the respective data and register an event if the latest value of the window is among the top 20\% of values in the rolling window. The resulting number of events registered for each country is roughly $500$. Following \cite{jalaldoust}, we assume that this data is observed within a time horizon of $T=400$ of the investigated exp-MHP, i.e. an instance of a short time horizon.

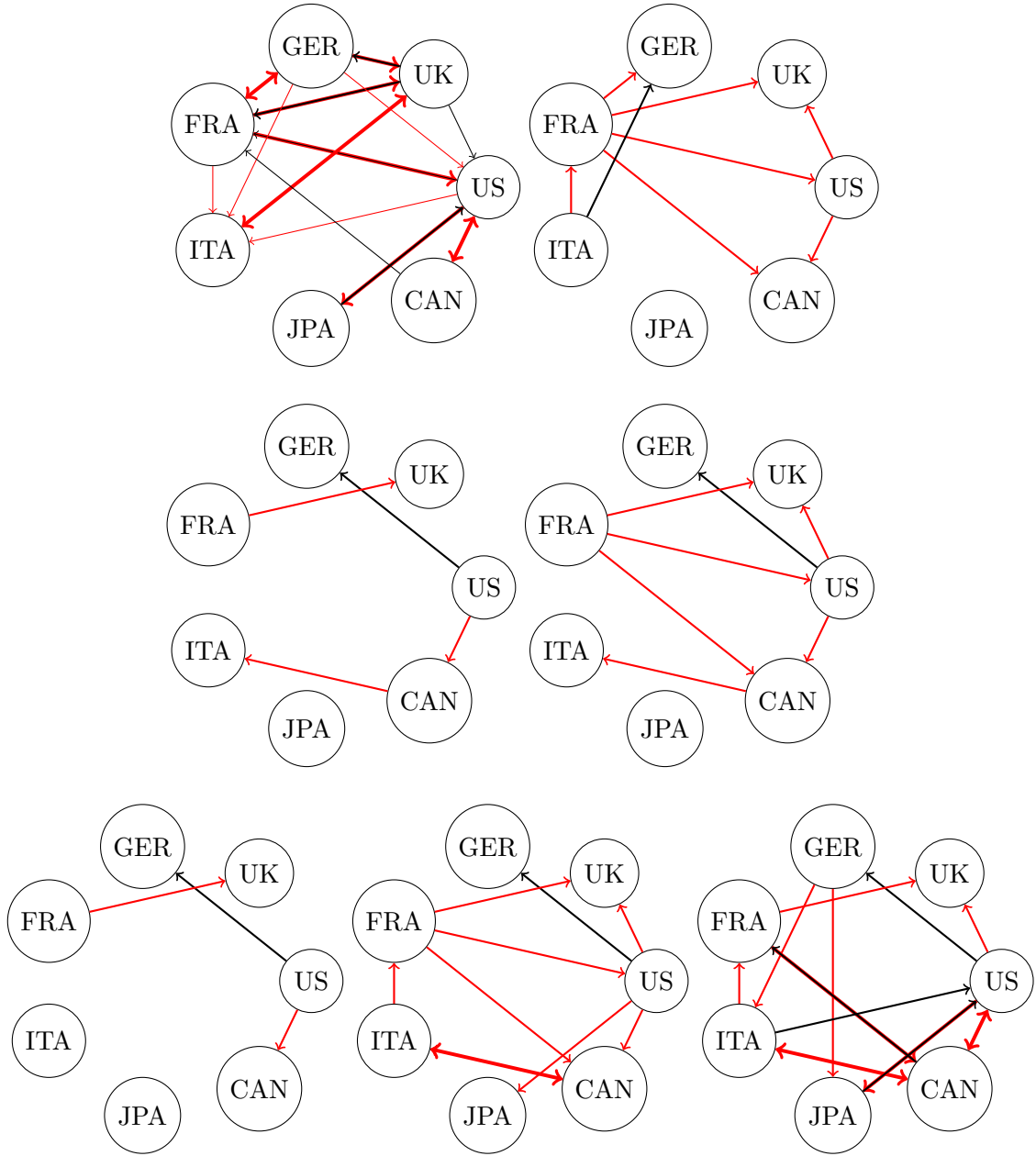
\begin{figure}[ht]
\centering
\begin{tikzpicture}
\def \n {7}
\def \radius {2.1cm}
\def \margin {8} 
\node[draw, circle] at ({360/7*0}:\radius ) (US) {US};
\node[draw, circle] at ({360/7*1}:\radius ) (UK) {UK};
\node[draw, circle] at ({360/7*2}:\radius ) (GER) {GER};
\node[draw, circle] at ({360/7*3}:\radius ) (FRA) {FRA};
\node[draw, circle] at ({360/7*4}:\radius ) (ITA) {ITA};
\node[draw, circle] at ({360/7*5}:\radius ) (JPA) {JPA};
\node[draw, circle] at ({360/7*6}:\radius ) (CAN) {CAN};
\draw[<->, thick,red,line width = 1.5] (US) -- (CAN);
\draw[<-, ] (US) -- (UK);
\draw[->, red,line width = 1.5] (FRA) -- (US);
\draw[->, ] (US) -- (FRA);
\draw[->, red] (US) -- (ITA);
\draw[->, thick,red,line width = 1.5] (US) -- (JPA);
\draw[->, thick] (JPA) -- (US);
\draw[->, red,line width = 1.5] (GER) -- (UK);
\draw[->, thick] (UK) -- (GER);
\draw[->, red,line width = 1.5] (FRA) -- (UK);
\draw[->, thick] (UK) -- (FRA);
\draw[<->, thick,red,line width = 1.5] (UK) -- (ITA);
\draw[->, red] (GER) -- (US);
\draw[<->, thick,red,line width = 1.5] (GER) -- (FRA);
\draw[->, red] (GER) -- (ITA);
\draw[->] (CAN) -- (FRA);
\draw[->, red] (FRA) -- (ITA);
\end{tikzpicture}
\begin{tikzpicture}
\def \n {7}
\def \radius {2.1cm}
\def \margin {8} 
  \node[draw, circle] at ({360/7*0}:\radius ) (US) {US};
\node[draw, circle] at ({360/7*1}:\radius ) (UK) {UK};
\node[draw, circle] at ({360/7*2}:\radius ) (GER) {GER};
\node[draw, circle] at ({360/7*3}:\radius ) (FRA) {FRA};
\node[draw, circle] at ({360/7*4}:\radius ) (ITA) {ITA};
\node[draw, circle] at ({360/7*5}:\radius ) (JPA) {JPA};
\node[draw, circle] at ({360/7*6}:\radius ) (CAN) {CAN};
\draw[->, thick,red] (FRA) -- (GER);
\draw[->, thick,red] (FRA) -- (UK);
\draw[->, thick,red] (FRA) -- (US);
\draw[->, thick,red] (FRA) -- (CAN);
\draw[->, thick] (ITA) -- (GER);
\draw[->, thick,red] (ITA) -- (FRA);
\draw[->, thick,red] (US) -- (UK);
\draw[->, thick,red] (US) -- (CAN);
\end{tikzpicture}
\\
\vspace{0.5cm}
\begin{tikzpicture}
\def \n {7}
\def \radius {2.1cm}
\def \margin {8} 
  \node[draw, circle] at ({360/7*0}:\radius ) (US) {US};
\node[draw, circle] at ({360/7*1}:\radius ) (UK) {UK};
\node[draw, circle] at ({360/7*2}:\radius ) (GER) {GER};
\node[draw, circle] at ({360/7*3}:\radius ) (FRA) {FRA};
\node[draw, circle] at ({360/7*4}:\radius ) (ITA) {ITA};
\node[draw, circle] at ({360/7*5}:\radius ) (JPA) {JPA};
\node[draw, circle] at ({360/7*6}:\radius ) (CAN) {CAN};
\draw[->, thick] (US) -- (GER);
\draw[->, thick,red] (US) -- (CAN);
\draw[->, thick,red] (FRA) -- (UK);
\draw[->, thick,red] (CAN) -- (ITA);
\end{tikzpicture}
\begin{tikzpicture}
\def \n {7}
\def \radius {2.1cm}
\def \margin {8} 
  \node[draw, circle] at ({360/7*0}:\radius ) (US) {US};
\node[draw, circle] at ({360/7*1}:\radius ) (UK) {UK};
\node[draw, circle] at ({360/7*2}:\radius ) (GER) {GER};
\node[draw, circle] at ({360/7*3}:\radius ) (FRA) {FRA};
\node[draw, circle] at ({360/7*4}:\radius ) (ITA) {ITA};
\node[draw, circle] at ({360/7*5}:\radius ) (JPA) {JPA};
\node[draw, circle] at ({360/7*6}:\radius ) (CAN) {CAN};
\draw[->, thick] (US) -- (GER);
\draw[->, thick,red] (US) -- (CAN);
\draw[->, thick,red] (US) -- (UK);
\draw[->, thick,red] (FRA) -- (UK);
\draw[->, thick,red] (FRA) -- (US);
\draw[->, thick,red] (FRA) -- (CAN);
\draw[->, thick,red] (CAN) -- (ITA);
\end{tikzpicture}
\vspace{0.5cm}
\\
\begin{tikzpicture}
\def \n {7}
\def \radius {2.0cm}
\def \margin {8} 
\node[draw, circle] at ({360/7*0}:\radius ) (US) {US};
\node[draw, circle] at ({360/7*1}:\radius ) (UK) {UK};
\node[draw, circle] at ({360/7*2}:\radius ) (GER) {GER};
\node[draw, circle] at ({360/7*3}:\radius ) (FRA) {FRA};
\node[draw, circle] at ({360/7*4}:\radius ) (ITA) {ITA};
\node[draw, circle] at ({360/7*5}:\radius ) (JPA) {JPA};
\node[draw, circle] at ({360/7*6}:\radius ) (CAN) {CAN};
\draw[->, thick] (US) -- (GER);
\draw[->, thick,red] (US) -- (CAN);
\draw[->, thick,red] (FRA) -- (UK);
\end{tikzpicture}
\begin{tikzpicture}
\def \n {7}
\def \radius {2.0cm}
\def \margin {8} 
  \node[draw, circle] at ({360/7*0}:\radius ) (US) {US};
\node[draw, circle] at ({360/7*1}:\radius ) (UK) {UK};
\node[draw, circle] at ({360/7*2}:\radius ) (GER) {GER};
\node[draw, circle] at ({360/7*3}:\radius ) (FRA) {FRA};
\node[draw, circle] at ({360/7*4}:\radius ) (ITA) {ITA};
\node[draw, circle] at ({360/7*5}:\radius ) (JPA) {JPA};
\node[draw, circle] at ({360/7*6}:\radius ) (CAN) {CAN};
\draw[->, thick] (US) -- (GER);
\draw[->, thick,red] (US) -- (CAN);
\draw[->, thick,red] (FRA) -- (UK);
\draw[->, thick,red] (US) -- (UK);
\draw[->, thick,red] (FRA) -- (US);
\draw[->, thick,red] (FRA) -- (CAN);
\draw[<->, thick,red,line width=1.5] (CAN) -- (ITA);
\draw[->, thick,red] (US) -- (JPA);
\draw[->, thick,red] (ITA) -- (FRA);
\draw[->, thick,red] (CAN) -- (ITA);
\end{tikzpicture}
\begin{tikzpicture}
\def \n {7}
\def \radius {2.0cm}
\def \margin {8} 
\node[draw, circle] at ({360/7*0}:\radius ) (US) {US};
\node[draw, circle] at ({360/7*1}:\radius ) (UK) {UK};
\node[draw, circle] at ({360/7*2}:\radius ) (GER) {GER};
\node[draw, circle] at ({360/7*3}:\radius ) (FRA) {FRA};
\node[draw, circle] at ({360/7*4}:\radius ) (ITA) {ITA};
\node[draw, circle] at ({360/7*5}:\radius ) (JPA) {JPA};
\node[draw, circle] at ({360/7*6}:\radius ) (CAN) {CAN};
\draw[->, thick,red] (US) -- (UK);
\draw[->, thick] (US) -- (GER);
\draw[->, thick,red,line width=1.5] (US) -- (JPA);
\draw[<->, thick,red,line width=1.5] (US) -- (CAN);
\draw[->, thick,red] (GER) -- (ITA);
\draw[->, thick,red] (GER) -- (JPA);
\draw[->, thick,red] (FRA) -- (UK);
\draw[->, thick,red,line width=1.5] (FRA) -- (CAN);
\draw[->, thick] (ITA) -- (US);
\draw[->, thick,red] (ITA) -- (FRA);
\draw[<->, thick,red,line width=1.5] (ITA) -- (CAN);
\draw[->, thick] (JPA) -- (US);
\draw[->, thick] (CAN) -- (FRA);
\end{tikzpicture}
\caption{Connectivity graph for ADM4 (top left panel), MDLH (top right panel), BIC (middle left panel), AIC (middle right panel), MMLH-e with $c=10^{-5}$ (bottom left panel), $c=0.3$ (bottom central panel), and $c=2.5$ (bottom right panel). The red connections are those that are in agreement with \cite{umar2022network}. 
}\label{fig:bondResults}
\end{figure}

Our study is conducted as follows: We compare MMLH with comparable methods designed for MHPs (in particular, MDLH, ADM4, BIC, and AIC) as well as with available expert knowledge. For MMLH, ADM4, BIC, and AIC, we prepare the data as described above and launch the respective causal discovery algorithm. For MDLH, we rely on the results reported in \cite{jalaldoust}. Moreover, as ``expert knowledge'' we consider the conclusions reported in \cite{umar2022network}, where the network connectedness between sovereign bond yield curve components of the G7 countries is discussed, without using MHPs as a basis. The connectivity graph derived from \cite{umar2022network} is shown in Figure~\ref{fig:umar}. The results of the investigated methods are presented in Figure~\ref{fig:bondResults}, with the graphs in the top panels corresponding to ADM4 (left) and MDLH (right) and the graphs in the middle panels corresponding to BIC (left) and AIC (right). In the bottom panels, we report the results for MMLH-e with $c=10^{-5}$ (left), $c=0.3$ (central), and $c=2.5$ (right). As expected, similar results are obtained for MMLH-u. For example, using $b=10^5$ we obtain the same graph as reported in the bottom left panel, and for $b=4$ the same as shown in the bottom central panel, except for the connection ``ITA to FRA'' not being present. Note that the red connections in the graphs visualized in Figure~\ref{fig:bondResults} are those that are in agreement with the expert knowledge graph of Figure~\ref{fig:umar}.

We observe that ADM4 yields the most connections in agreement with \cite{umar2022network}, however it also suggests $6$ connections that are not present in Figure \ref{fig:umar}. As expected from the synthetic experiments, for small $c$ (resp. large $b$) (i.e., when we have a strong penalty on structures $\bgamma_i \in \bGamma_i=\{0,1\}^p$ with many non-zero entries), MMLH performs similar to BIC, which only suggests one connection not present in Figure \ref{fig:umar} (``US to GER''). Increasing $c$ to $0.3$ (resp. decreasing $b$ to $4$) adds further connections to the graph (bottom central panel), of which all are reported in \cite{umar2022network}. In particular, for $c=0.3$ MMLH-e yields $9$ connections present in Figure \ref{fig:umar} (and a single one that is not reported there). In comparison, MDLH and AIC only capture $7$ (resp. $6$) connections of Figure \ref{fig:umar} (and also report one connection that is not reported there). Thus, the proposed MMLH algorithm not only outperforms the classical BIC and AIC methods, but also reports more connections that are in agreement with the literature than the related state-of-the-art MDLH method. Note also that $6$ out of $8$ connections detected by MDLH and all connections obtained by AIC are in agreement with MMLH-e for $c=0.3$. 

Moreover, we observe that many outgoing connections from France are captured well by ADM4, MDLH, AIC and MMLH. Note, however, that for MMLH-e the connection ``FRA to US'' disappears when $c$ is increased from $0.3$ (bottom central panel) to $2.5$ (bottom right panel).  This may be explained by the newly discovered edges now dominating the connection ``FRA to US'', in the sense that the chosen structure shown in the bottom right panel yields a slightly smaller value of criterion \eqref{eq:MML87_full_expMHP_k} than the same structure including the connection ``FRA to US''. A similar effect is observed for MMLH-u. Further, the outgoing connections from Germany are only (partially) captured by ADM4. However, this improves for MMLH under large values of $c$ (resp. small values of $b$). In particular, using $c=2.5$ and $b=0.25$, we obtain the connections ``GER to ITA'' and ``GER to JPA''.
We also observe that Japan is isolated from the other countries, except under ADM4 and MMLH, which report an ingoing connection from the US, in agreement with \cite{umar2022network}. Only MMLH-e (for $c=2.5$) reports a second ingoing connection (from Germany), which is also present in Figure \ref{fig:umar}. These observations also suggest a solid performance of the proposed MMLH procedure.

\section{Conclusion and Discussion}\label{section:conclusion}

In this paper, we estimated Granger causal relations between components of multivariate Hawkes processes with exponential decay kernels. These relations are described by a connectivity graph, which can be estimated from observations of the process.
We approached this problem by proposing an optimization criterion and a model selection algorithm MMLH based on the minimum message length principle.

In contrast to other model selection algorithms, MMLH incorporates prior distributions of the underlying model parameters.
While classical model selection criteria (e.g. BIC) are designed to penalize models with a lot of parameters (i.e. structures with many non-zero entries) and thus do not take into account other descriptions of the model, MMLH offers more flexibility in terms of structure-related penalty. This may be particularly beneficial, if some a-priori expert knowledge on the structure of the underlying graph exists. Given the fact that all model parameters to be estimated are non-negative, we investigated both a uniform prior and an exponential prior and observed a similar performance for both of them in all our experiments.

We conducted synthetic experiments in which we compared the proposed algorithm to other related classical and state-of-art methods, focusing on short time horizons. In the considered sparse graph scenarios, MMLH achieved the highest F1 scores among all comparison methods. Concerning the investigated mid-dense scenario, MMLH showed an F1 score comparable to MLE-ms, MLE-thr, AIC and ADM4. However, the performance of MMLH improved with increasing time horizons, the only rivals being the state-of-art MDLH and the classical AIC. The superior F1 score of MMLH on sparse connectivity graphs may be explained by the fact that the minimum message length principle prefers short encodings of the model (i.e.  sparse graphs) over longer encodings (i.e. non-sparse graphs) together with a short description of the data using the model.

Finally, we illustrated the proposed method on G7 sovereign bond data and compared the inferred causal connections to those of MDLH, ADM4, BIC, and AIC. We demonstrated that the connectivity graphs obtained via MMLH (with three different parameterizations of the prior function) are in agreement with the expert knowledge extracted from the literature.

As a possible future work one may investigate connectivity graphs in multivariate Hawkes processes with other kernels or intensities given by non-linear functional relationships (e.g., ReLU or sigmoid functions). Another research direction is a modification of the algorithm which allows to increase the performance on non-sparse structures. Moreover, one may study if/how the presented method benefits from considering directed acyclic graph structures. For recent approaches in this regard, we refer to  \cite{zheng2018dags}, \cite{zhang2022truncated}, and \cite{wei2023causal}.

\acks{The authors would like to thank the reviewers for their constructive comments which helped to improve the manuscript. Their support is highly appreciated. K. H.-S.  acknowledges a partial support by the Austrian Science Foundation (FWF) project I 5113-N,     
and by the Czech Academy of Sciences, Praemium Academiae awarded to M.
Palu\v s. A part of this paper was written while
I.T. was member of the Institute of Stochastics, Johannes Kepler University Linz, 4040 Linz,
Austria. During this time, I.T. was supported by the Austrian Science Fund (FWF): W1214-N15,
project DK14.  
}

\vskip 0.2in
\bibliography{sample}

\end{document}